\documentclass[twoside,11pt]{article}

\usepackage[left=2.5cm,right=2.5cm,top=2.3cm,bottom=2.3cm]{geometry}

\usepackage{amssymb,amsmath,amsfonts}
\usepackage{graphicx}
\usepackage{latexsym}
\usepackage{color}
\usepackage{algorithmic}
\usepackage{algorithm}
\usepackage{multirow}
\usepackage{natbib}
\usepackage{url}
\usepackage{fancyhdr} 

\newtheorem{theorem}{Theorem}
\newtheorem{proof}{Proof}
\newtheorem{corollary}{Corollary}

\newcommand{\SEQL}{{\sc seql }}
\newcommand{\SEQLLR}{{\sc seql-lr }}
\newcommand{\SEQLSVM}{{\sc seql-svm }}

\newcommand{\SSSKSVM}{{\sc sssk-svm }}
\newcommand{\SVM}{{\sc svm }}

\title{Bounded Coordinate-Descent for Biological Sequence Classification in High Dimensional Predictor Space}

\author{Georgiana Ifrim, Carsten Wiuf \\
        {\{ifrim, wiuf\}@birc.au.dk} \\
       {\it Bioinformatics Research Centre (BiRC)} \\
       {\it C.F. M\o llers All\'e 8} \\
	{\it DK-8000 Aarhus C, Denmark}}

\date{}

\begin{document}

\maketitle

\begin{abstract}
We present a framework for discriminative sequence classification where the learner works directly in the high dimensional predictor space of all subsequences 
in the training set. This is possible by employing a new coordinate-descent algorithm coupled with bounding the magnitude of the gradient for selecting discriminative subsequences fast. We characterize the loss functions for which our generic learning algorithm can be applied and present concrete implementations for logistic regression (binomial log-likelihood loss) and support vector machines (squared hinge loss). Application of our algorithm to protein remote homology detection and remote fold recognition results in performance comparable to that of state-of-the-art methods (e.g., kernel support vector machines). Unlike state-of-the-art classifiers, the resulting  classification models are simply lists of weighted discriminative subsequences and can thus be interpreted and related to the biological problem.
\end{abstract}

\section{Introduction}

Many problems in biology today require accurate computational prediction of properties. For example, the primary DNA sequence is a main determinant of functional and structural protein properties, yet little is known about this relationship and we must therefore turn to computational prediction for advancing our understanding. Likewise, accurate gene and motif prediction is of crucial importance for the annotation of recently sequenced genomes.
To achieve this goal we need machine learning techniques that are fast, highly scalable and preferably treat feature selection as an integral part of the learning algorithm. The latter requirement means that neither time nor expertise is invested in pre-processing the original data (e.g., for defining features) and no potentially hard to validate assumptions are made about data distribution. Recent advances in developing efficient machine learning tools such as fast logistic regression \citep{genkin:bbr07,lin:nlr08} and support vector machines \citep{joachims:svmperf06,hsieh:svm08} enable learning of classifiers in very large predictor spaces, thus reducing the need for pre-processing the data. Nevertheless, most of these techniques are typically designed to exploit the sparsity of the training set (i.e., many features occur sparsely in the training data) which holds for many applications, such as text categorization \citep{lin:nlr08,hsieh:svm08}. This assumption unfortunately does not hold for biological sequences, where many shorter features often occur very frequently.

We present an efficient coordinate-descent algorithm for optimizing regularized fitting of classification loss in high dimensional predictor space. Our approach does not rely on feature sparsity assumptions. In our framework, the feature space is spanned by all subsequences present in the training set. Furthermore, the features can have a flexible number of wildcard matches, which allows us to model complex biological processes such as substitutions, insertions and deletions. The optimization proceeds coordinate-wise by iteratively selecting the feature with maximum (absolute) gradient value, following the Gauss-Southwell rule \citep{luenberger:optimbook84}. In order to select the best features fast, we provide bounds on the gradient value of any subsequence based on its prefix. 
This drastically reduces the search space.
We discuss the tightness of the proposed bounds as well as the convergence of the algorithm.

Our learning technique is applicable to both unregularized and regularized loss functions. In this paper we show bounds for the more complex case of \emph {elastic-net} regularized loss \citep{zhou:elasticnet05}. By adding an explicit elastic-net penalty (a convex combination of $l1$ and $l2$ regularizers) to the loss function, we allow the user to directly trade-off $l1$-regularization (encouraging model sparsity) for $l2$-regularization (correcting for correlations) \citep{hastie:slbook03, zhou:elasticnet05, friedman:regpath10}. As can be observed in our experiments this positively affects the prediction quality.

We present applications of our learning algorithm to protein remote homology detection and fold recognition.
In order to compare to previously published results, we work with standard protein benchmarks. For homology detection we use SCOP1.59 \citep{jaakkola:ismb99, leslie:nips02, leslie:jmlr04, kuksa:biokdd08}, and for fold recognition we use the challenging dataset of \citet{ding:bioinf01}. The classification problems associated with these datasets are hard and the training data is rather small scale (2,800 and 300 sequences respectively). In order to further analyse the scalability of our technique, we present a large scale experiment on the latest version of the Silva-LSUPARC102 database (150,000 unique sequences). We compare our algorithm to state-of-the-art sequence classifiers, support vector machines with spectrum kernel \citep{leslie:jmlr04}, mismatch kernel \citep{leslie:nips02, leslie:bioinf04}, and the recent sparse spatial sample kernel \citep{kuksa:biokdd08, kuksa:nips08}.
Besides being fast and highly accurate, our classification models are easily interpretable, an important advantage for the bioinformatics and medical communities.

The remainder of this paper proceeds as follows. In Section~\ref{related-work} we discuss related work. In Section~\ref{learning-algorithm} we present our learning algorithm. In Section~\ref{experiments} we discuss the experimental setup. In Section~\ref{results} we discuss the empirical results and we conclude in Section~\ref{conclusion}.

\section{Related Work}
\label{related-work}

Working in high dimensional predictor space and regularizing is statistically preferable to a two-step procedure of first reducing the dimension, then fitting a model in the reduced space \citep{rosset:boosting04}.
Recently, efficient regularized learning algorithms for logistic regression and support vector machines (SVM) were proposed for fitting classifiers in high dimensional predictor space \citep{lin:nlr08,hsieh:svm08, perkins:grafting03}.
These algorithms are still at least linear in the number of features. For applications in which the number of features is much larger than the number of training samples, this poses a challenge for both running time and memory \citep{ifrim:kdd08}. This is the case when we try to use all (unrestricted-length) subsequences in the training set as features.
If the features are restricted to $k$-mers, for fixed and reasonably small $k$ (i.e., $k=3$, or $k=5$) and the features occur sparsely in the training instances (such as in applications related to text categorization), existing classification algorithms may still perform well \citep{ifrim:phdthesis09}. However, this is not the case for applications focusing on biological sequence classification where the feature space is dense, i.e., many subsequences occur in many of the training sequences.

Recent work on efficiently computing string kernels for SVM \citep{leslie:nips02, leslie:jmlr04, kuksa:biokdd08} addressed some of the the computational challenges typically associated with this type of techniques. Nevertheless, the kernels proposed still restrict the set of features to subsequences of certain length or format (e.g., allowing a certain number of mismatches) in order to deal with the computational aspects associated with large feature spaces.
In this paper we work with the unrestricted space of all subsequences in the training set and allow wildcard matches.

A popular class of models used for sequence classification is the generative classifier family, such as profile Hidden Markov Models \citep{eddy:bioinf98}. Generative models only learn from positive training examples, while discriminative techniques (such as SVM and those proposed in this paper) directly focus on separating the positive and negative training examples. Previous work has shown that discriminative approaches outperform generative approaches for sequence classification \citep{cheng:bioinf06, leslie:nips02, kuang:bioinf04}.

In this paper we propose learning a linear classifier directly in a high dimensional feature space rather than using an implicit mapping via a string kernel as in kernel-SVM. The computational trick in our algorithm is to use coordinate descent coupled with bounding the search for the best coordinate. Our generic algorithm can be implemented for a range of classification loss functions, such as exponential loss, binomial log-likelihood loss of logistic regression, guassian loss and the squared hinge-loss of SVM \citep{hastie:slbook03}. This work is an extension of the study of \citet{ifrim:kdd08} which focused on a simpler version of this algorithm for unregularized logistic regression applied to text categorization.

\section{Bounded Coordinate-Descent for Sequence Classification}
\label{learning-algorithm}

In this section we present our generic learning algorithm and its concrete implementation for logistic regression and support vector machines.

\subsection{Preliminaries and Basic Notation}

We first introduce the theoretical framework and some basic notation.
Assume we have a training set of instance-label pairs $\{x_i,y_i\}_{i=1}^{N}$, with $y_i \in \{-1,+1\}$.
The training instances $x_i$ are sequences, e.g., biological sequences $x_i = AGTCAACTGGAA....$, text sequences $x_i=ABCD...$, 
sequences of rankings $x_i=A < B < C < D < \dots$, sequences of conjunctions of properties $x_i=A \cap B \cap C \cap D \cap \dots$.
Let $d$ be the number of distinct subsequences in the feature space. 
We formally represent the training sequences as binary vectors in the space of all subsequences of the training set:
$x_i = (x_{i1}, \dots x_{ij}, \dots x_{id})^T, \; x_{ij} \in \{0,1\}, \; i=\overline{1,N}$.
Let $\beta=(\beta_{1}, \dots, \beta_{j}, \dots, \beta_{d})$ be a parameter vector (defining a linear classifier).

The goal is to learn a mapping, also called classification model, $f: X \rightarrow \{-1,+1\}$ from the given training set such that given a new sample $x \in X$, we can predict a label $y \in \{-1, +1\}$.

Let 
\begin{equation}
\label{genobjfct}
L(\beta)=\sum_{i=1}^{N} \xi(y_i,x_i,\beta) + C R_{\alpha}(\beta)
\end{equation} be the regularized classification loss criterion.
Here, $\xi(y_i,x_i,\beta)$ is a classification loss function, $C$ is a constant and $R_{\alpha}(\beta)$ is a regularizer. 
Learning a classification model is achieved by finding the parameter vector $\beta$, that minimizes the regularized classification loss on the training set
\begin{equation}
\hat \beta = \textrm{argmin}_{\beta} L(\beta).\nonumber
\end{equation}
In Equations (\ref{exploss})--(\ref{gaussloss}) we list a few examples of commonly used classification loss functions. 
\begin{equation}
\label{exploss}
Exponential \; loss: \xi(y_i,x_i,\beta) = e^{-y_i \beta^t x_i}
\end{equation}
\begin{equation}
\label{lrloss}
Binomial \; loglikelihood \; loss: \xi(y_i,x_i,\beta) = \log(1 + e^{-y_i \beta^t x_i})
\end{equation}
\begin{equation}
\label{hingeloss}
Squared \; hinge \; loss: \xi(y_i,x_i,\beta) = \max(1 - y_i  \beta^t  x_i, 0)^2
\end{equation}
\begin{equation}
\label{gaussloss}
Gaussian \; loss\footnote{\text{$\Phi$ is the cdf of the standard normal distribution.}}: \xi(y_i,x_i,\beta) = \log \frac{1}{\Phi(y_i  \beta^t  x_i)}
\end{equation}

The penalty weight $C \geq 0$ controls the amount of regularization of the parameter vector $\beta$. A large $C$ means more penalty on the $\beta$ parameters. The regularization term can take various forms. The most common regularizers are the $l1$ which penalizes the sum of absolute values of $\beta_j$, and the $l2$, which penalizes the sum of squared $\beta_j$ coeficients (Equation (\ref{l1reg})). In our work $R_{\alpha}(\beta)$ is the elastic-net regularizer \citep{zhou:elasticnet05} (defined in Equation (\ref{enreg})), which is a compromise between the $l2$ $(\alpha=0)$ and the $l1$ $(\alpha=1)$ regularizers \citep{friedman:regpath10}.
\begin{equation}
\label{l1reg}
l1 = \sum_{j=1}^{d} |\beta_j|, \;\;\;\;\;\;
l2 = \frac{1}{2}\sum_{j=1}^{d} \beta_j^2 
\end{equation}
\begin{equation}
\label{enreg}
R_{\alpha}(\beta) = \alpha \sum_{j=1}^{d} |\beta_j|  + (1-\alpha) \frac{1}{2} \sum_{j=1}^{d} \beta_j^2 
\end{equation}

Depending on the type of regularization, increasing the weight $C$ of the penalty term results in many zero $\beta_j$ (for $l1$-regularization) or shrinking the coefficients $\beta_j$ of correlated predictors (for $l2$-regularization). The elastic-net regularizer allows balancing the two effects.

\subsection{Generic Bounded Coordinate-Descent Algorithm}

In this section we describe our learning algorithm and characterize the properties of loss functions to which it can be applied.
Assume the following properties for the loss function:
\begin{itemize}
\item[1.]  $\xi$ depends on $y_i, \; x_i$ and $\beta$ only through the classification margin $m_i=y_i \beta^t x_i$. Note that $x_i$ is correctly classified if the margin $m_i$ is positive and the higher the margin the higher the classification confidence.
We write $\xi(y,x,\beta)=\xi(m)$.
\item[2.] $\xi$ is a monotone decreasing function of the margin: $\xi'(m) \leq 0$.
\item[3.] $\xi$ is strictly convex and twice continuously differentiable.
\end{itemize}
The gradient of $L(\beta)$ (defined in Equation (\ref{genobjfct})) with respect to a coordinate $\beta_j$ is then
\begin{equation}
\frac{\partial L}{\partial \beta_j}(\beta)= \sum_{i=1}^{N}  y_i  x_{ij}  \xi'(y_i \beta^t x_i) + C [ \alpha \; \textrm{sign}(\beta_{j}) + (1-\alpha)  \beta_j]  \nonumber
\end{equation}
or using the margin notation
\begin{equation}
\frac{\partial L}{\partial \beta_j}(\beta)= \sum_{i=1}^{N}  y_i  x_{ij}  \xi'(m_i) + C   R'_{\alpha}(\beta_{j}).\nonumber
\end{equation}
If $R_{\alpha}(\beta)$ is not differentiable in $\beta=0$ we use the left-right derivatives.
We proceed minimizing $L(\beta)$ by coordinate-wise gradient descent in the feature space of all subsequences.
In each iteration we update only the coordinate corresponding to the subsequence with the largest gradient magnitude, 
following the Gauss-Southwell rule:
\begin{equation}
\label{bestj}
j = \textrm{argmax}_{l} \left|\frac{\partial L}{\partial \beta_{l}}(\beta)\right| \nonumber
\end{equation} 
This results in a greedy advance towards the optimum of the objective function, without having to explicitly materialize the full space of subsequences.
All we need is an efficient way to find the best coordinate (i.e., feature) in each iteration.
We summarize the steps of our generic coordinate-descent algorithm in Algorithm \ref{alg:gencoorddesc}.
\begin{algorithm}[h!]
\caption{Generic coordinate-descent algorithm}
\label{alg:gencoorddesc}
\begin{enumerate}
\item Set $\beta^{(0)} = 0$
\item For t=1:T
	\begin{itemize}
	\item[(a)] Find $j_t = \textrm{argmax}_j \left|\frac{\partial L}{\partial \beta_{j_{t}}}(\beta^{(t-1)})\right|$ (using Theorem \ref{theorem:pruning})
	\item [(b)] Use line search to set the step length $\epsilon_t$ for coordinate $j_{t}$
	\item [(c)] Set $\beta^{(t)}_{j_{t}} = \beta^{(t-1)}_{j_{t}} - \epsilon_t  \frac{\partial L}{\partial \beta_{j_{t}}}(\beta^{(t-1)})$ and $\beta^{(t)}_{k} = \beta^{(t-1)}_{k}, k \neq j_t$
	\end{itemize}
\end{enumerate}
\end{algorithm}

The core part of the algorithm is a search routine (line (a) in Algorithm \ref{alg:gencoorddesc}) that returns fast the best feature in each iteration. The search procedure relies on bounding the gradient value of any subsequence early on, by only looking at its prefix. This looks difficult at a first glance since the prefix may be a poor feature, while its extension can be highly discriminative.
The intuition behind our bound is based on two observations: the gradient is influenced by class-wise frequency and the subsequence space has structure.
More concretely, the gradient of a given feature is characterized by how many times this feature appears in the positive class versus the negative class. A feature has high gradient value if it is frequent in one class and not in the other. If we look at the gradient of a given feature class-wise, it is influenced only by the frequency of the given feature in the given class, and in turn, the frequency of the feature in that class is bounded by the frequency of its prefix. Thus, class-wise we can bound the gradient of a given subsequence based on its prefix, and then we can build a global bound from the class-wise bounds.
Using these bounds we can discard large parts of the search space during the search for the best subsequence, making the whole search process efficient, both time-wise (fast running time) and space-wise (low memory).
 
Let $j$ be a coordinate corresponding to a given subsequence $s_j$, and $l$ be a coordinate corresponding to a super sequence of $s_j$, $s_{l}$, i.e. $s_j$ is a prefix of $s_{l}$. We write $s_j \in x_i$ to denote $x_{ij} \neq 0$.

The following theorem gives a tight upper bound on the gradient value of any subsequence.
\begin{theorem}
\label{theorem:pruning}
For any loss function $\xi$ satisfying properties 1-2 and any subsequence $s_{l} \supseteq s_j$, $j=  1, \dots, d$, 
\begin{eqnarray}
\label{th:bound}
 \left|\frac{\partial L}{\partial \beta_{l}}(\beta)\right| \leq & \max & \left\{\left|\sum_{\{i| \mathbf{s_j \in x_i}, y_i=+1\}}  \xi'(m_i) + C  R'_{\alpha}(\beta_{l}) \right|\right.,\\
&& \left.\left|\sum_{\{i| \mathbf{s_j \in x_i}, y_i=-1\}}  -\xi'(m_i) + C  R'_{\alpha}(\beta_{l}) \right|\right\}.\nonumber
\end{eqnarray}

\begin{proof}
We split the analysis in two parts, the first part focuses on deriving a bound within the negative class, and the second part focuses on the positive class.
Recall that $\xi'(m) \leq 0$. Then, 
\begin{eqnarray}
\label{gradbound}
\frac{\partial L}{\partial \beta_{l}}(\beta) & = & \sum_{\{i| s_{l} \in x_i\}}  y_i  x_{il}  \xi'(m_i) + C  R'_{\alpha}(\beta_{l})  \\\nonumber
& \leq & \sum_{\{i| s_{l} \in x_i, y_i = -1\}} y_i  x_{il}  \xi'(m_i) + C  R'_{\alpha}(\beta_{l})  \\\nonumber
& \leq & \sum_{\{i| s_{j} \in x_i, y_i = -1\}}  y_i  x_{ij}  \xi'(m_i) + C  R'_{\alpha}(\beta_{l}) \\\nonumber
& = & \sum_{\{i| s_{j} \in x_i, y_i = -1\}}  -\xi'(m_i) + C  R'_{\alpha}(\beta_{l}) \nonumber
\end{eqnarray}
The last inequality in Equation~(\ref{gradbound}) holds because of the anti-monotonicity property $\{i| s_{l} \in x_i, y_i=-1\} \subseteq \{i|s_j \in x_i, y_i=-1\}$.
Similarly, we can show for the positive class that 
\begin{eqnarray}
\frac{\partial L}{\partial \beta_{l}}(\beta) & \geq & \sum_{\{i| s_{j} \in x_i, y_i = +1\}}  \xi'(m_i) + C  R'_{\alpha}(\beta_{l}) \nonumber
\end{eqnarray}
Thus we obtain the lower and upper bounds
\begin{equation}
\sum_{\{i| s_{j} \in x_i, y_i = +1\}}  \xi'(m_i) + C  R'_{\alpha}(\beta_{l})  
\leq 
\frac{\partial L}{\partial \beta_{l}}(\beta) \\\nonumber
\leq  
\sum_{\{i| s_{j} \in x_i, y_i = -1\}}  -\xi'(m_i) + C  R'_{\alpha}(\beta_{l}). \nonumber
\end{equation}
We are interested in an upper bound on the absolute magnitude of the gradient at a given coordinate, thus we write more conveniently
\begin{eqnarray}
\label{upperbound}
\left|\frac{\partial L}{\partial \beta_{l}}(\beta)\right|  \leq & \max & \left\{\left|\sum_{\{i| s_{j} \in x_i, y_i = +1\}}  \xi'(m_i) + C  R'_{\alpha}(\beta_{l}) \right|\right., \\\nonumber
& & \left.\left|\sum_{\{i| s_{j} \in x_i, y_i = -1\}}  -\xi'(m_i) + C  R'_{\alpha}(\beta_{l}) \right|\right\}. \nonumber
\end{eqnarray}
\end{proof}
\end{theorem}
The main property on which Theorem \ref{theorem:pruning} relies on (last inequality in Equation~(\ref{gradbound})) is the anti-monotonicity: the prefix frequency is higher than that of its extension subsequence. This means that the same bound holds for \emph{any} weight $x_j$ as long as $x_j \geq x_l$. This observation is useful for example if we want to integrate prior biological knowledge about the target problem, such as the fact that some amino acids in protein sequences are more useful than others. We could encode this information by giving weights $w_a \in (0,1)$ to individual amino acids and still have the property that $x_j=\Pi_a w_a \geq x_l$.  

As we can observe from Equation (\ref{upperbound}), without explicit regularization of the loss function, the upper bound on the gradient of a subsequence $s_l$ solely depends on the frequency of its prefix $s_j$. This means that we can decide not to expand the $s_j$ prefix further, without inspecting the longer feature $s_l$.
If we use a regularizer, we have the term $\beta_{l}$ in the bound which depends on the longer subsequence, rather than only the prefix. Thus, we have to take care to compute the correct bound whenever $\beta_l \neq 0$. Since at start all $\beta_{j}$ are zero, during iterations we only need to check those features that have non-zero $\beta_{l}$ because they were selected in the model in previous iterations. In practice, as supported by our experiments, this part is fairly fast due to the sparsity of the final model learned by this algorithm. 

The bound presented in Theorem \ref{theorem:pruning} is tight since we can construct examples for which the inequality is an equality. Consider the simple case of a training set with positive examples of the type {\em AAAAAAA}, and negative examples of the type {\em BBBBBB}. Whenever the subsequence set of occurrences is the same as that of its prefix, the inequality becomes equality.

The iterates generated by Algorithm \ref{alg:gencoorddesc} converge to the optimal solution of the objective function given in Equation (\ref{genobjfct}).
Based on results from \citet{luo:convergence92} characterizing the convergence of iterates generated by coordinate-descent using the Gauss-Southwell rule we have:
\begin{theorem}
Let $\beta^{(t)}$ be a sequence of iterates generated by Algorithm \ref{alg:gencoorddesc}. 
Then $\beta^{(t)}$ converges to the optimal solution of (\ref{genobjfct}).
\end{theorem}

In the next subsections we describe the specific bounds used to design learning algorithms in the style of Algorithm 1 for logistic regression and support vector machines.
We choose these popular classifiers for concrete implementations because their loss functions satisfy the properties 1-3 and they are margin maximizing \citep{rosset:boosting04}. The latter property typically translates into good generalization ability in practice.

\subsection{Logistic Regression}

Let 
\begin{equation}
L(\beta) = \sum_{i=1}^{N}\log(1 + e^{-y_i \beta^t x_i}) + C R_{\alpha}(\beta)\nonumber
\end{equation}
be the elastic-net-regularized binomial log-likelihood loss \citep{hastie:slbook03}. 
The gradient of $L(\beta)$ with respect to a coordinate $j$ at a given parameter vector $\beta$ is:
$$\frac{\partial L}{\partial \beta_j}(\beta)=\sum_{i=1}^{N}  y_i  x_{ij}  \underbrace{\left(\frac{-1}{1 + e^{-y_i  \beta^t  x_i}}\right)}_{\xi'(m_i)} + C  R'_{\alpha}(\beta_{l}) $$
\begin{corollary}
\label{cor:pruninglr}
For binomial log-likelihood loss and any subsequence $s_{l} \supseteq s_j$, $j= 1, \dots, d$, 
\begin{eqnarray}
\left|\frac{\partial L}{\partial \beta_{l}}(\beta)\right| \leq \max \left\{\left|\sum_{\{i| s_{j} \in x_i, y_i = +1\}}  \frac{-1}{1 + e^{-\beta^t  x_i}} + C  R'_{\alpha}(\beta_{l}) \right|\right., \\ \nonumber
\left.\left|\sum_{\{i| s_{j} \in x_i, y_i = -1\}}  \frac{1}{1 + e^{\beta^t  x_i}} + C  R'_{\alpha}(\beta_{l}) \right|\right\}.\nonumber
\end{eqnarray}
\end{corollary}

\subsection{Support Vector Machines}

Let
\begin{equation}
L(\beta) = \sum_{i=1}^{N}\max(1 - y_i  \beta^t  x_i, 0)^{2} + C  R_{\alpha}(\beta) \nonumber
\end{equation}
be the elastic-net-regularized squared hinge loss \citep{chang:l2svm08}. 
We can rewrite $L(\beta)$ in the equivalent form
\begin{equation}
L(\beta) = \sum_{\{i| 1 - y_i \beta^t  x_i > 0\}} (1 - y_i  \beta^t  x_i)^{2} + C  R_{\alpha}(\beta)\nonumber
\end{equation}
The gradient of $L(\beta)$ with respect to a coordinate $j$ at a given parameter vector $\beta$ is:
$$\frac{\partial L}{\partial \beta_j}(\beta)=\sum_{\{i| 1 - y_i \beta^t  x_i > 0\}}  y_i  x_{ij}  \underbrace{2(y_i  \beta^t  x_i - 1)}_{\xi'(m_i)} + C  R'_{\alpha}(\beta_{l})  $$

\begin{corollary}
\label{cor:pruningsvm}
For squared hinge loss and any subsequence $s_{l} \supseteq s_j$, $j=1, \dots, d$, 
\begin{eqnarray}
\left|\frac{\partial L}{\partial \beta_{l}}(\beta)\right| \leq \max \left\{\left|\sum_{\{i| s_{j} \in x_i, 1 - \beta^t  x_i > 0, y_i = +1\}}  2(\beta^t  x_i - 1) + C  R'_{\alpha}(\beta_{l}) \right|\right., \\
\left.\left|\sum_{\{i| s_{j} \in x_i, 1 + \beta^t  x_i > 0, y_i = -1\}}  2(1 + \beta^t  x_i) + C  R'_{\alpha}(\beta_{l}) \right|\right\}.\nonumber
\end{eqnarray}
\end{corollary}

\subsection{Algorithmic Details}

In this section we give some details on using the bound of Theorem \ref{theorem:pruning} to prune the search space.

Before starting the optimization iterations we build an inverted index on all the distinct unigrams in the training set, i.e., a list of document ids and positions of occurrence for each unigram. In each iteration we start the process of searching for the best feature from the level of unigrams. 

We implement two strategies of prefix-expansion. The breadth-first-search (BFS) expands all unigrams to bi-grams, then all bi-grams to tri-grams, etc.
For each unigram, we compute the gradient value, we keep track of the best gradient/feature seen so far and we compare the bound to the current best gradient. If the bound is lower than the current best gradient value, no subsequence starting with this unigram can improve the current optimum and we prune this part of the search space (i.e., we discard this prefix). Then we move on to the next level of expansion, and repeat this process.
The depth-first-search (DFS) strategy expands the unigram to its longest subsequence until the pruning condition is met, and it then backtracks to the longest valid prefix and re-attempts to expand. Depending on the sequence tokenization used, word-level or character-level, the two strategies have different benefits.
For word-level tokenization, such as in text categorization, short subsequences are more likely to be useful than long ones (e.g., phrases of 2-3 words are typically good discriminators), and therefore BFS may be a better expansion strategy.
For character-level tokenization, such as for biological sequences, longer subsequences tend to be more useful than shorter ones (e.g., subsequences of length 3 or less may occur in all the sequences), so DFS is a better expansion choice.
For the experiments in this paper we have used depth-first-search. 
In order to keep track of the occurrences of all active prefixes (e.g., subsequences that could not be pruned using Theorem \ref{theorem:pruning}), we 
expand the inverted index on-demand. The inverted index does not grow excessively due to the effectiveness of the bound.
After selecting the feature with the best gradient in a given iteration we do a line search to compute the final feature weight.
We use a binary search strategy for the line search: we double an initial small step size until breaking the monotonicity of the loss function (i.e., the loss function stops decreasing), after which we narrow down the search by halving the step until reaching a required precision. We stop the optimization iterations based on a convergence thereshold on the loss function.

The worst-case complexity of each iteration is $O(d + N)$, where $d$ is the size of the feature space and $N$ is the number of training samples. However, in practice our algorithm drastically prunes the search space and is thus very fast.
For example, for the classification task {\em Scorpion-toxin like} (presented in detail in Section \ref{results}) we have observed the following behaviour. There are 23 distinct unigrams in the training set (22 amino acids and we add the wildcard as an additional unigram). Assuming we restrict the $k$-mer size to $k=10$ (only for the scope of this example, in our experiments $k$ is unrestricted), the total number of features is then $d=23^{10}\sim10^{14}$. For finding the best feature in the first iteration our algorithm checks the pruning bound 1,113 times, and prunes the space 1,104 (out of 1,113) times. The algorithm stops after 54 iterations. In the last iteration, it checks the bound 11,336 times and prunes the space 11,239 times. Thus using the bound in Theorem \ref{theorem:pruning}, we prune the search space to less than {\em a billionth of a percent} for this particular task.

For simplifying the implementation we currently prune the space using a prefix-expansion strategy. Theorem \ref{theorem:pruning} however can be used for pruning all subsequences that contain a given subsequence, not necessarily in the start position as a prefix. This observation could be used to further speedup the training process.

\section{Experiments}
\label{experiments}
We implement the generic coordinate-descent algorithm in our machine learning tool {\bf \sc seql} ({\bf SEQ}uence {\bf L}earner), available from \url{http://www.birc.au.dk//~ifrim/seql} under the GNU GPL open source license.
For now, \SEQL implements the two elastic-net-regularized classifiers presented in the previous sections: logistic regression and support vector machines.
In this section we describe the datasets, the techniques compared and the methodology for designing experiments.

\subsection{Datasets}

In order to compare our results to the state-of-the-art we report directly the results published in those respective papers and perform experiments on the same benchmarks.

For {\bf protein remote homology detection} we use SCOP1.59 \citep{leslie:nips02, leslie:jmlr04, weston:bioinf05, kuksa:biokdd08}. This is an expert-curated database of protein domains organized heirarchically into folds, superfamilies and families. Protein sequences belonging to different families but the same superfamily are considered to be remote homologs in SCOP.
This dataset contains 2,862 labeled sequences organized into 54 binary classification problems simulating homology detection by predicting the super-family using a hold-one-family-out strategy. No pair of sequences shares more than 95\% identity. The positive training sets are quite small and the learning task is challenging. For more details on this benchmark see \citet{jaakkola:ismb99, leslie:jmlr04,weston:bioinf05}.
 
For {\bf protein remote fold recognition} we use the benchmark published by \citet{ding:bioinf01}. This dataset consists of sequences from 27 folds divided into two independent sets such that the training and test sequences share less than 35\% sequence identities and within the training set, no sequences share more than 40\% sequence identities. There are 311 training sequences and 383 test sequences.

In order to analyze the {\bf scalability} of our method in a large scale experiment, we have downloaded the latest {\bf ribosomal RNA} (rRNA) database Silva-LSUParc\footnote{Silva-LSUParc database: http://www.arb-silva.de/documentation/background/release-102/.} \citep{pruesse:silva-lsuparc102}, version 102 (released in February 2010.) The LSUParc102 database contains 180,344 rRNA sequences.
After removing duplicate sequences we obtain a dataset of 150,780 unique sequences organized in 3 (one-vs-all) binary classification tasks according to the Bacteria, Archaea and Eukarya domains. The distribution by domain is dominated by Eukarya with 141,601 sequences, followed by Bacteria with 8,967 and Archaea with 212 sequences.

All datasets are available from \url{http://www.birc.au.dk//~ifrim/seql/data}.

\subsection{Techniques Compared}

Previous studies have shown that discriminative approaches for sequence classification (such as kernel-SVM) outperform generative approaches (such as profile HMM) by a large margin \citep{cheng:bioinf06, kuang:bioinf04, leslie:nips02, liao:recomb02}.

\begin{figure}[h!]
  \centering
\vspace*{-0.5cm}
\includegraphics[scale=0.5]{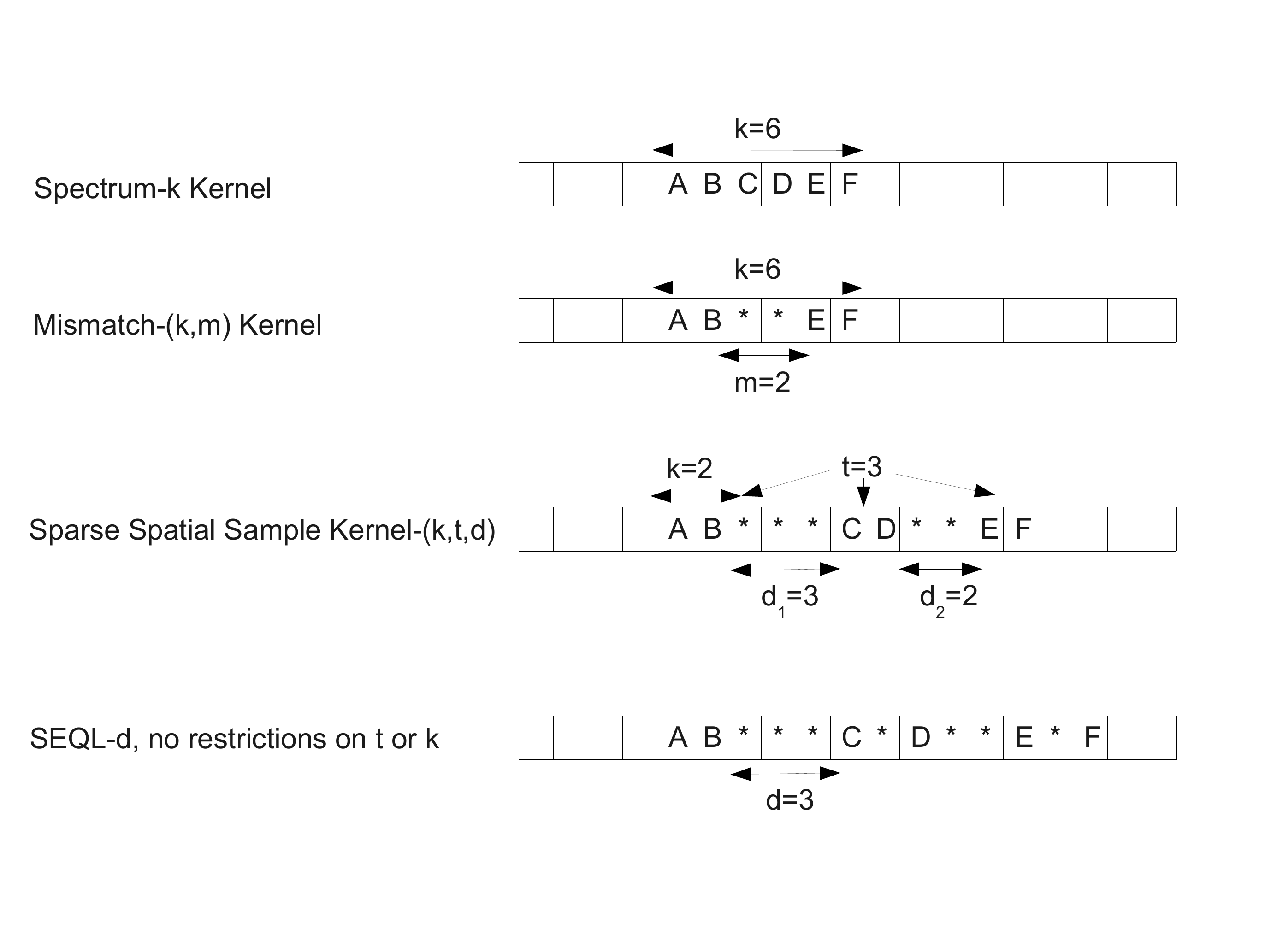}
\vspace*{-0.5cm}
\caption{Features employed by state-of-the-art kernels versus our method {\sc SEQL}. {\bf Spectrum-k Kernel}, {\bf k} is the length of features. {\bf Mismatch-(k,m) Kernel}, {\bf k} is the length of features, {\bf m} is the maximum number of mismatches. {\bf Sparse Spatial Sample-(k,t,d) Kernel}, {\bf k}  is the probe size, {\bf t} is the number of probes and {\bf d} is the number of maximum allowed positions between the probes. {\bf {\sc SEQL}-d}, {\bf d} is the maximum number of consecutive wildcards.} 
  \label{fig:features}
\end{figure}

We compare our algorithms to the latest techniques based on sequence kernels for SVM: the spectrum kernel, the mismatch kernel \citep{leslie:nips02, leslie:jmlr04, lodhi:jmlr02}, and the recent sparse spatial sample kernel (SSSK) \citep{kuksa:nips08, kuksa:biokdd08}.
All these methods aim at computing similarity for all pairs of sequences (the kernel matrix) in a particular feature space. Due to computational challenges, these techniques typically restrict the length and expressive power of the subsequences used as features.
For example, the spectrum-$k$ kernel \citep{leslie:psb02} implicitly compares sequences in the space of all $k$-mers, where the length $k$ of subsequence-features is a parameter of the model.
The mismatch kernel \citep{leslie:nips02} generalizes the spectrum-$k$ kernel by allowing up to $m$ mismatches or substitutions to accomodate mutations.
The sparse spatial sample kernel (SSSK) further generalizes the mismatch kernel by sampling the sequences at different resolutions and comparing the resulting spectra \citep{kuksa:biokdd08}. SSSK has 3 parameters, $(k,t,d)$, where $k$ is the probe size, $t$ is the number of probes and $d$ is the number of {\em maximum} allowed positions between the probes. Our learning algorithm uses all (unrestricted-length) subsequences in the training set as features. Furthermore, we also allow mismatches or so called wildcard matches, by a parameter that controls the {\em maximum} number of consecutive wildcards allowed. This allows us to model complex biological processes such as substitutions, insertions and deletions.
Figure \ref{fig:features} gives examples for the types of features implicitly used by the above described kernels, as compared to the features used by our technique.

One advantage of kernel techniques is their ability to integrate unlabeled data during kernel computation to relax labeled data requirements. However, computing all-pair similarities for large datasets (resulting from the addition of unlabeled sequences) remains computationally very challenging for kernel methods both time and memory-wise. For example, we couldn't apply any of the above mentioned kernel methods on our large dataset (150,000 sequences) since this would require more than 90GByte memory. 

For completion we also report the results of a technique published by \citet{liao:recomb02} which applies SVM using an empirical kernel map based on pairwise Smith-Waterman sequence alignment scores (SVM-pairwise in Table\ref{table:scop159-auc50}).

We use SEQL-LR and SEQL-SVM to denote the logistic regression and support vector machines implementations of our generic learning algorithm.

\subsection{Methodology}

Previous studies on biological sequence classification have extensively used the AUC (area under the ROC curve) and AUC50 for evaluation \citep{fawcett:rocgraphs04, leslie:jmlr04, sonego:rocbioinf08}. AUC describes the ranking of prediction scores rather then being dependent on a fixed classification threshold. The ROC curve is obtained by plotting the fraction of true positives versus the fraction of false positives for a binary classifier as its discrimination threshold is varied \citep{fawcett:rocgraphs04}. 
A common aggregate measure is to report the area under the ROC curve (AUC) where an area of 1 represents a perfect ranking of all positives above all negatives and an area close to 0.5 represents a random classifier.
The AUC50 focuses on top ranked examples and is defined as the normalized area under the ROC curve computed for up to 50 true negatives \citep{gribskov:cc96}. It is typically used to evaluate classifiers on datasets where the number of positives is much lower than the number of negatives.
While for a complete test set the AUC is betwen 0.5 and 1, the AUC values for truncated top lists are between 0 and 1. The shorter the top list is, the smaller the AUC values \citep{sonego:rocbioinf08}.
In some studies instead of AUC or AUC50 only the balanced-error-rate is reported. To compare our results to state-of-the-art, we also show the balanced-error (Equation (\ref{ber})) which measures the average of the errors on each class for a fixed classification threshold. It can equivalently be interpreted in terms of Specificity and Sensitivity \citep{levner:balanced-error-rate06}.
\begin{equation}
\label{ber}
BER = \frac{1}{2}\left(\underbrace{\frac{FN}{TP+FN}}_{1-\text{Sensitivity}}+\underbrace{\frac{FP}{FP+TN}}_{1-\text{Specificity}}\right)
\end{equation}

The benchmarks for remote homology detection and fold recognition have pre-defined training and test splits. We use the same data for experiments for all methods compared, and for the prior techniques {\em we report directly the results published in those respective papers}.
For the large scale experiment on ribosomal RNA, we show 5-fold cross-validation results on the full dataset. Since the prior techniques required more than the available memory, for the large dataset we only report results using our method.

All experiments were run on a Linux machine with 2.5GByte memory and 2.8GHz Intel CPU.

\section{Results and Discussion}
\label{results}

In this section we present the results of applying \SEQLLR and \SEQLSVM to protein remote homology detection and fold recognition.
Furthermore, to evaluate the scalability of our approach, we present a large-scale experiment on the latest release of the Silva-LSUParc database \citep{pruesse:silva-lsuparc102}.

\subsection{Comparison to State-of-the-Art}

We first compare the performance of \SEQLLR and \SEQLSVM to that of kernel-SVM techniques on two protein classification tasks.

\subsubsection{Protein remote homology detection}

In Table~\ref{table:scop159-auc50} we show results on the SCOP1.59 benchmark for all compared methods. The numbers in brackets next to the name of each method refer to the explicit parameters of these methods as discussed in the previous section, e.g., length of subsequences used as features or maximum number of wildcards. The results are averaged over all 54 binary classification tasks  which correspond to superfamilies.
\begin{table}[h!]
\begin{center}
\small
\begin{tabular}{|l|c|c|c|}
\hline
Method			&	AUC  	& 	AUC50	& BER\\\hline
\sc spectrum-(2)	&	0.8581	&	0.3583	& - \\\hline		
\sc spectrum-(3)	&	0.8723	&	0.4037	& - \\\hline
\sc mismatch-(5,1) 	& 	0.8749 	&	0.4167	& -\\\hline
\sc sssk(1,2,5)		& 	0.8901 	& 	0.4629	& -\\\hline
\sc sssk(1,3,3)		& 	0.9148	& 	0.5118	& - \\\hline
{\sc svm}-pairwise   	& 	0.8930 	&	0.4340	& -\\\hline
\sc seql-lr(5)		&	\bf 0.9106	& \bf	0.5185	& \bf 0.4418 \\\hline
\sc seql-svm(5)		& 	\bf 0.9214 	& \bf 0.5213	&  \bf 0.4383 \\\hline
\end{tabular}
\end{center}
\caption{{\bf Remote homology detection on the Scop1.59 dataset.} The average AUC, AUC50 and BER scores over the 54 target superfamilies.
Results for kernel-SVM methods cited directly from \citet{leslie:jmlr04, kuksa:biokdd08}.}
\label{table:scop159-auc50}
\end{table}

For \SEQLLR and \SEQLSVM we set the maximum allowed number of wildcards to 5. We varied the regularization parameters for the elastic net penalty. We show the most accurate results over the set of parameter values tried.
We observe that even if in this benchmark the positive training sets are quite small (minimum number of samples is 2 and maximum is 95) our techniques outperform state-of-the-art methods. We believe this is due to the additional flexibility of features used for learning classifiers. Our approach does not restrict the length of features, but rather lets the data drive the classifier's decisions as to which subsequences are most discriminative. We also note that \SEQLSVM has slightly better AUC and AUC50 than \SEQLLR.
Learning \SEQL classifiers on this dataset took a few seconds per topic and about 80MByte memory. 
The computational requirements of \SEQL are mainly influenced by the maximum allowed number of wildcards (set to 5 in this experiment).

\begin{figure}[h!]
  \centering
\includegraphics[scale=0.75]{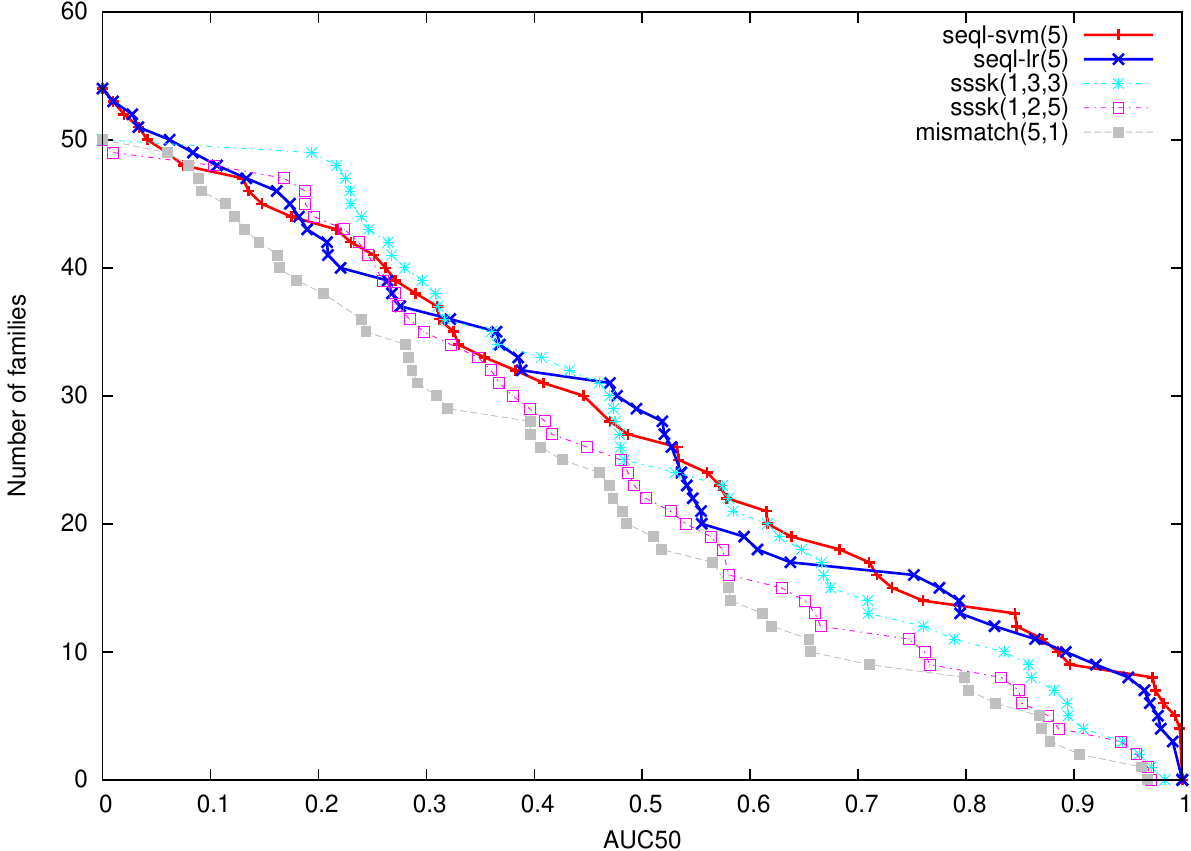}
\caption{{\bf Comparison of mismatch-SVM, SSSK-SVM, SEQL-LR and SEQL-SVM on SCOP1.59.} The graph plots the total number of superfamilies for which a given method exceeds an AUC50 score threshold.}
  \label{fig:scop159-auc50}
\end{figure}

Figure~\ref{fig:scop159-auc50} shows a graphic comparison of some of the methods (for which we had access to published per class AUC50 scores) by plotting the total number of superfamilies above an AUC50 score threshold as a function of the threshold value. 
The threshold is changed so as to generate the next AUC50 value in the ranked list of AUC50 scores of the respective method.
The numerical integral of this cumulative AUC50 curve is the arithmetic average of the AUC50 values shown in Table~\ref{table:scop159-auc50} \citep{sonego:rocbioinf08}. SVM with sparse-sample-spatial kernel ({\sc sssk-svm}) behaves best among the prior techniques.
We observe that \SEQLLR and \SEQLSVM are comparable to {\sc sssk-svm} with some gain on families where \SSSKSVM has either low or good quality (AUC50 zero or above 0.4). Even though the features employed by \SSSKSVM are quite flexible (see Figure~\ref{fig:features}), we still gain some performance by imposing even less restrictions on features.

\begin{figure}[h!]
  \centering
\includegraphics[scale=0.6]{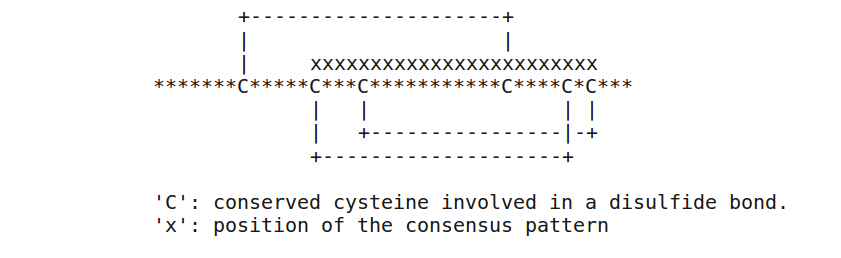}
\caption{Schematic representation of the {\em short-chain scorpion toxins} family from the PROSITE database.}
  \label{fig:prosite}
\end{figure}

\citet{kuksa:biokdd08} analyzed the biological relevance of features learned by \SSSKSVM taking the {\em Scorpion toxin-like} superfamily as an example. This classification task has 16 members of this superfamily as positive training examples, 1067 non-members as negative examples and the {\em short-chain scorpion toxin} family as a test case. {\sc sssk-svm}(1,2,5) achieved an AUC50 of 0.7661 for this task  \citep{kuksa:biokdd08}. In their work \citet{kuksa:biokdd08} present the schematic representation of the {\em short-chain scorpion toxin} family obtained from PROSITE \citep{hulo:prosite06}. The PROSITE database consists of a large collection of manually-curated biologically meaningful signatures that are described as patterns or profiles \citep{hulo:prosite06}.
In Figure~\ref{fig:prosite} we show the PROSITE representation. 
The scheme shows that these type of toxins contain six conserved cysteines (C) residues involved in disulfide bonds. PROSITE also lists a consensus pattern present in all the members of this family shown as well in Figure~\ref{fig:prosite} (marked with 'X').
We focus here on the same superfamily and look at the features learned by {\sc seql-svm}(5). 
\begin{table}[h!]
\begin{center}
\small
\begin{tabular}{ccccc}
\begin{tabular}{|l|c|}
\hline
\multicolumn{2}{|c|}{Positive Features}\\\hline\hline
Weight & Feature \\\hline
0.1551	& SG*C \\
0.1518	& GYC \\
0.1388	& N**C***C \\
0.0885	& C*C \\
0.0816	& C*****A****C \\
0.0657	& G***G*C \\
0.0654 	& C***C***G \\
0.0484 	& C***C \\
0.0479 	& C*****C***C \\
0.0394 	& K**C \\\hline
\end{tabular}
&&&&
\begin{tabular}{|l|c|}
\hline
\multicolumn{2}{|c|}{Negative Features}\\\hline\hline
Weight & Feature \\\hline
-0.1449 & G \\
-0.0783 & CP \\
-0.0754 & T \\
-0.0541 & R****L \\
-0.0502 & S*R \\
-0.0462 & C***T \\
-0.0428 & A**L \\
-0.0405 & S*****K \\
-0.0400 & C**G \\
-0.0382 & E \\\hline
\end{tabular}
\end{tabular}
\end{center}
\caption{Top-10 positive and negative \SEQLSVM features on the SCOP.1.59 Scorpion-toxin-like superfamily.}
\label{table:scop159-feat}
\end{table}
In Table~\ref{table:scop159-feat} we show the top-10 positive and negative features selected by \SEQLSVM on this superfamily. 

{\sc seql-svm}(5) has an AUC50 of 0.8847 on this task. \citet{kuksa:biokdd08} analyzed the support vectors of their method and observed that the feature {\sc C***C} had the highest weight. Similarly, {\sc seql-svm}(5) selects the feature {\sc C***C} in top-10, but ranks features such as {\sc N*C***C} and {\sc C*C} higher, which boosts its AUC50 score. In order to understand the effect of these features on the AUC50 score, in Table~\ref{table:scop159-top5} we take a closer look at the training/test statistics of some of them. 
\begin{table}[h!]
\begin{center}
\small
\begin{tabular}{|l||l|l|l|l|}
\hline
                      	& \multicolumn{2}{c}{Training (superfamily)}			& \multicolumn{2}{|c|}{Test (family)} \\\hline
Feature			& Positive		& Negative		& Positive	& Negative \\\hline
1. SG*C                 & 62.5\% (10/16)	& 1.5\% (17/1067)     	& 13\%  (3/23)	& 1.2\% (21/1661)\\
2. GYC                  & 62.5\% (10/16)       	& 0.5\% (6/1067) 	& 0.4\% (1/23)  & 0.09\% (15/1661)\\
3. N**C***C             & 62.5\% (10/16)      	& 0.02\% (3/1067)       & 0.4\% (1/23)  & 0.03\% (6/1661)\\
4. C*C                  & 100\%  (16/16)       	& 5.7\% (61/1067)       & 100\% (23/23) & 4.4\% (74/1661) \\
8. C***C                & 100\% (16/16)        	& 6.3\% (68/1067)       & 100\% (23/23) & 6.1\% (102/1661)\\\hline
\end{tabular}
\end{center}
\caption{Quality of the top ranked positive {\sc seql-svm}(5) features for the Scorpion-toxin-like superfamily.}
\label{table:scop159-top5}
\end{table}
\begin{figure}[ht!]
  \centering
\includegraphics[scale=0.65]{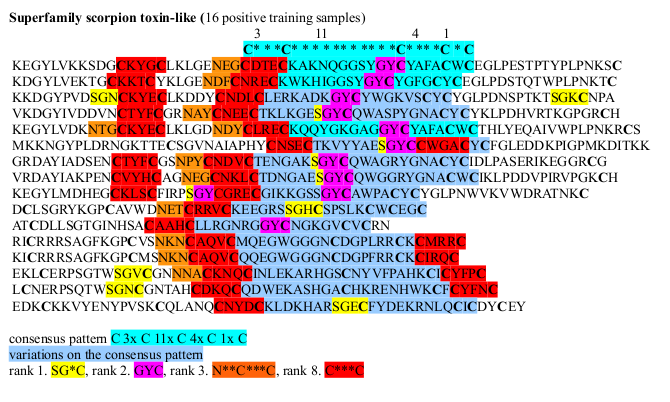}
\caption{Position of top \SEQLSVM features relative to the PROSITE consensus pattern.}
  \label{fig:scop159-feat-consensus}
\end{figure}
We show the percentage of occurrence followed by absolute numbers in the brackets.
The notation 10/16 means that the respective feature occurs in 10 samples out of 16. Since the dataset is very unbalanced, percentages are a better indicator of feature quality. We observe that although of high quality, the feature {\sc c***c} also occurs in quite a few negatives and it is less useful than {\sc c*c} for example.
Note that the final ranking does not necessarily correspond to the order of selection during optimization iterations. For example the feature {\sc c*c} was selected before {\sc sg*c} in the optimization iterations, but due to the specific loss function, the repeated selection of {\sc sg*c} lead to a slightly higher weight for this feature in the final model.

In order to asses the importance of these automatically selected features relative to the manually-curated pattern, we show in Figure~\ref{fig:scop159-feat-consensus} their exact location in the training sequences. We note that most of these features are part of the consensus pattern. 

%
%
%
%
%
%
%

\subsubsection{Protein remote fold recognition}

In this section we show an application of \SEQLLR and \SEQLSVM to protein remote fold recognition on the dataset published by \citet{ding:bioinf01}. They proposed a technique based on building an SVM classifier using features deemed biologically relevant for this problem, such as percentage composition of the 20 amino acids, predicted secondary structure, polarity.
In Table~\ref{table:dd} we show results comparing the technique of \citet{ding:bioinf01}, denoted by {\sc svm(d\&d)}, with published results of kernel-SVM methods and with our \SEQLLR and \SEQLSVM algorithms. Recall that lower BER means better classification quality.
\begin{table}[h!]
\begin{center}
\small
\begin{tabular}{|l|c|c|c|}
\hline
Method		&	AUC  	& 	AUC50	& BER\\\hline
\SVM(D\&D) 	&	-	&	-	& 0.5650\\\hline
\sc mismatch(5,1)   &	-	&	-	& 0.5322\\\hline
\sc sssk(1,2,5)    &	-	&	-	& 0.4619\\\hline
\sc sssk(1,3,3)    &	-	&	-	& 0.4499\\\hline
\sc seql-lr(5)	&	\bf 0.7705	& \bf	0.3519	& \bf 0.4407 \\\hline
\sc seql-svm(5)	& \bf 	0.7848 	& \bf 0.3317	& \bf 0.4289 \\\hline
\end{tabular}
\end{center}
\caption{{\bf Remote fold recognition on the D\&D dataset.} The average AUC, AUC50 and BER scores over the 27 target folds. Results for SVM(D\&D) and for the other kernel-SVM methods cited directly from \citet{ding:bioinf01, kuksa:biokdd08}.}
\label{table:dd}
\end{table}

Even if we don't use any domain-specific knowledge in this experiment, we observe that our techniques have better classification quality than the method of \citet{ding:bioinf01} which relies on deep biological insight. We believe this advantage results from treating feature selection as a part of the learning algorithm and allowing for a large degree of flexibility in the features.
\SEQLLR and \SEQLSVM outperform the state-of-the-art techniques in terms of balanced-error-rate on this benchmark.

\subsection{Large-Scale Experiment}

In this section we analyze the scalability of our method on a ribosomal RNA (rRNA) domain-prediction task. 

\subsection{Ribosomal RNA Domain-Prediction}

The dataset used for this task contains 150,780 unique rRNA sequences.
It is estimated that based on the new capacity for cheap and rapid sequencing there is a steady flow of about 10,000 rRNA sequences per month into the public sequence databases \citep{pruesse:silva-lsuparc102}. Furthermore, many sequences are derived from cultivation independent biodiversity surveys, which rely on rapid pattern- or clone-based approaches that often generate partial rRNA sequences \citep{pruesse:silva-lsuparc102}.

We show the average results of \SEQLLR and \SEQLSVM over 5-fold cross-validation splits for fixed parameter values.
We set the amount of regularization $C=1.0$, we balance $l1$ and $l2$ regularization by setting $\alpha=0.5$ and allow no wildcards by setting the maximum number of consecutive wildcards $d=0$.

In Table~\ref{table:silva-lsuparc102} we show classification quality, running time and memory resources required by \SEQLLR and \SEQLSVM.
\begin{table}[h!]
\begin{center}
\small
\begin{tabular}{|l|c|c|c|c|c|}
\hline
Method		&	AUC  	& 	AUC50	& BER		& Time		& Memory\\\hline
\SEQLLR		&	0.9981	& 	0.9722	& 0.0088 	& 30 min	& 2GByte\\\hline
\SEQLSVM	&  	0.9987 	&  	0.9653	& 0.0083 	& 20 min	& 2GByte\\\hline
\end{tabular}
\end{center}
\caption{Ribosomal RNA domain-prediction on Silva-LSUParc102.}
\label{table:silva-lsuparc102}
\end{table}
We note that both {\sc seql}-based techniques have high classification quality. Additionally, they only take about half-an-hour running-time and a reasonable 2GByte memory instead of 90GByte required by a kernel matrix computation.
Instead of building sequence classifiers by costly multiple sequence alignment as currently done for sequence retrieval in this database, we could directly learn domain {\sc seql}-classifiers from the original input sequences in the respective domains.

\section{Conclusion}
\label{conclusion}

In this paper we have presented a new learning algorithm for sequence classification in high dimensional predictor space. The algorithm has at its core a coordinate-wise gradient descent strategy coupled with bounding the magnitude of the gradient to retrieve high quality features fast. We have characterized the loss functions to which our generic learning algorithm can be applied and have presented concrete implementations for logistic regression and support vector machines.
Based on applications to protein remote homology detection and fold recognition, as well as large-scale domain-prediction for ribosomal RNA, we have observed our techniques are comparable to state-of-the-art in terms of classification quality. In addition, the techniques presented are highly scalable and the resulting classification models can easily be interpreted and connected to biologically relevant facts.\\\\
{\bf Acknowledgements} {We thank Deepak Ajwani for helpful comments on the manuscript. This work is supported by the Danish Cancer Society.}

\appendix
\section*{Appendix A. Parameter Options for SEQL}

In this appendix we give a list of \SEQL parameters the user can set in order to influence the outcome classification model. \SEQL is available from
\texttt{http://www.birc.au.dk/$\sim$ifrim/seql}.

\begin{description}

\item[$\langle$-o objective$\rangle$]  Objective function. Choice between logistic regression and support vector machines. By default set to logistic regression.
\item[$\langle$-g maxgap$\rangle$] Maximum number of consecutive gaps or wildcards allowed in a feature, e.g., {\sc a**b}, is a feature of size 4 with any 2 characters from the input alphabet in the middle. By default set to 0.
\item[$\langle$-C regularizer\_value$\rangle$] Value of the regularization parameter. By default set to 1.
\item[$\langle$-a alpha$\rangle$] Weight of l1 vs l2 regularizer for the elastic-net penalty. By default set to 0.5.
\item[$\langle$-l minpat$\rangle$] Threshold on the minimum length of any feature. By default set to 1. 
\item[$\langle$-L maxpat$\rangle$] Threshold on the maximum length of any feature. By default the maximum length is unrestricted, i.e., at most as long as the longest sequence in the training set.
\item[$\langle$-m minsup$\rangle$] Threshold on the minimum support of features, i.e., number of sequences containing a given feature. By default set to 1.
\item[$\langle$-n token\_type$\rangle$] Word or character-level token. Words are delimited by white spaces. By default set to character-level tokens.
\item[$\langle$-r traversal\_strategy$\rangle$] Breadth First Search or Depth First Search traversal of the search tree. By default set to DFS.
\item[$\langle$-c convergence\_threshold$\rangle$] Stopping threshold based on change in aggregated score predictions. By default set to 0.005.
\item[$\langle$-T maxitr$\rangle$]  Number of optimization iterations. By default set to the maximum between 5,000 and the number of iterations resulting by using a convergence threshold on the aggregated change in score predictions.
\item[$\langle$-v verbosity$\rangle$] Amount of printed detail about the training of the classifier. By default set to 1 (light profiling information).

\end{description}

\vskip 0.2in

\bibliography{seql-arxiv2010}
\bibliographystyle{plainnat}

\end{document}